\documentclass{article}

\PassOptionsToPackage{numbers, compress}{natbib}

\usepackage[final]{neurips_2020}

\usepackage[english]{babel}
\usepackage[utf8]{inputenc}

\usepackage[nottoc]{tocbibind}

\usepackage[utf8]{inputenc} 
\usepackage[T1]{fontenc}    
\usepackage{hyperref}       
\usepackage{url}            
\usepackage{booktabs}       
\usepackage{amsfonts}       
\usepackage{nicefrac}       
\usepackage{microtype}      
\usepackage{amsmath}
\usepackage[dvipsnames]{xcolor}
\usepackage{algorithm2e}
\usepackage{graphicx}
\usepackage{textcomp}

\title{GPU Accelerated Exhaustive Search for Optimal Ensemble of Black-Box Optimization Algorithms}

\author{%
  Jiwei~Liu\\
  Nvidia\\
  RAPIDS\\
  Pittsburgh, PA, USA \\
  \texttt{jiweil@nvidia.com} \\
  \And
  Bojan~Tunguz\\
  Nvidia\\
  RAPIDS\\
  Greencastle, IN, USA \\
  \texttt{btunguz@nvidia.com} \\
  \And
  Gilberto~Titericz\\
  Nvidia\\
  RAPIDS\\
  Curitiba, Brazil \\
  \texttt{gtitericz@nvidia.com} \\
}

\begin{document}

\maketitle

\begin{abstract}
  Black-box optimization is essential for tuning complex machine learning algorithms which are easier to experiment with than to understand. In this paper, we show that a simple ensemble of black-box optimization algorithms can outperform any single one of them. However, searching for such an optimal ensemble requires a large number of experiments. We propose a Multi-GPU-optimized framework to accelerate a brute force search for the optimal ensemble of black-box optimization algorithms by running many experiments in parallel. The lightweight optimizations are performed by CPU while expensive model training and evaluations are assigned to GPUs. We evaluate 15 optimizers by training 2.7 million models and running 541,440 optimizations. On a DGX-1, the search time is reduced from more than 10 days on two 20-core CPUs to less than 24 hours on 8-GPUs. With the optimal ensemble found by GPU-accelerated exhaustive search, we won the $2^{nd}$ place of NeurIPS 2020 black-box optimization challenge \footnote{Source code: https://github.com/daxiongshu/rapids-ai-BBO-2nd-place-solution}. 
\end{abstract}

\section{Introduction}
Black-box optimization (BBO) has become the state-of-the-art methodology for parameter tuning of complex machine learning models \cite{golovin2017google,eriksson2019scalable}. The optimization process is considered black-box because the details of the underlying machine learning models, datasets and the objective functions are hidden from the optimizer. The optimizer finds best hyper-parameters by experimenting with the machine learning models and observing the performance scores \cite{bayesmark}. Popular BBO algorithms include random search \cite{bergstra2012random} and Bayesian optimization \cite{shahriari2015taking}. Random search is proven to be more effective than brute-force grid search \cite{bergstra2012random}. Bayesian optimization (BO), on the other hand, utilizes probabilistic models to find better hyper-parameters and it outperforms random search consistently \cite{shahriari2015taking}. Many advanced BO algorithms have been proposed to improve its scalability \cite{hernandez2017parallel,eriksson2019scalable}. Popular BBO libraries such as Optuna \cite{optuna-2019} and Hyperopt \cite{bergstra2013making} adopt the Tree of parzen estimators (TPE) \cite{bergstra2011algorithms} algorithm and efficient pruning techniques. 

As advances made in BBO improve accuracy, efficiency and usability, they also become increasingly complicated and opaque to users, just like another black box. The NeurIPS BBO challenge \cite{blackbox} provides a great opportunity to study and improve them. We made the following observations from what we learned:

\begin{itemize}
\item BBO algorithms excel in different machine learning models, datasets and objective functions.
\item The overall execution time is dominant by model evaluation, which could cost 100x time of the actual optimization. 
\end{itemize}

These insights inspire us to treat BBO algorithms as black-boxes and search for an ensemble of BBO algorithms that outperforms the best single BBO algorithm. A simple exhaustive search is proven to be effective and it is enabled by accelerating massive parallel model evaluations on GPUs. Specifically we made the following contributions:
\begin{itemize}
\item An ensemble algorithm that allows multiple BBO algorithms to collectively make suggestions and share observations, within the same time budget as a single BBO algorithm.
\item A multi-GPU optimized exhaustive search framework to find BBO candidates for the optimal ensemble.
\item A suite of GPU-optimized cuML \cite{raschka2020machine} models including scikit-learn counterparts, MLPs and Xgboost \cite{Chen:2016:XST:2939672.2939785} are added to the Bayesmark toolkit to accelerate single model evaluation.
\item A comprehensive evaluation and empirical analysis of both individual and ensemble optimization algorithms.
\end{itemize}

\section{Overview of Bayesmark and Provided Optimizers}
Bayesmark \cite{bayesmark} is the framework for the BBO challenge, which has scikit-learn \cite{scikit-learn} (sklearn) models and datasets built-in to evaluate BBO algorithms. The Bayesmark provides 6 optimization algorithms, 9 machine learning models and 6 datasets, which are summarized in table \ref{Table1}.

Machine learning models are from scikit-learn toolkit \cite{scikit-learn} and each model has two variants for classification and regression, respectively. In the competition, each function is optimized in $N\_STEP=16$ iterations with batch size of $N\_BATCH=8$ per iteration. The optimizer for Bayesmark implements an suggest-observe interface as shown in Algorithm 1 (Left). Each iteration the optimizer suggests N\_BATCH new guesses for evaluating the function. Once the evaluation is done, the scores are passed back as observations to update the optimizers. The function to be optimized is simply the cross validation score of a machine learning model on a dataset with a loss function.

\begin{table}
  \caption{Bayesmark Overview}
  \label{Table1}
  \centering
  \begin{tabular}{ll}
    \midrule
    Optimizers & hyperopt \cite{bergstra2013making}, 
    opentuner \cite{ansel:pact:2014}, pysot \cite{eriksson2019pysot}, skopt \cite{skopt}, turbo \cite{eriksson2019scalable}     \\
    Models \cite{scikit-learn}  & DT, MLP-adam, MLP-sgd, RF, SVM, ada, kNN, lasso, linear \\
    Dataset \cite{scikit-learn,datametrics} & breast, digits, iris, wine, boston, diabetes \\
    Metrics \cite{scikit-learn,datametrics} & nll, acc, mse, mae \\
  \end{tabular}
\end{table}

\section{Motivation Study}
We study the performance of given optimizers with $N\_STEP=16$, $N\_BATCH=8$ and repeat 3 times ($N\_REPEAT=3$). Experiments are run using the default script \cite{runlocal} for all sklearn models, datasets and metrics. 

Table \ref{Table2} shows the optimization timing breakdown averaged per iteration and function for each optimizer. The time budget for optimization, which is the limit for the total time of “suggest” and “observe”, cannot exceed 40 seconds at each iteration. It is apparent that the time budget is more than enough to run multiple optimizations, suggesting an opportunity for an ensemble of optimizers.

Table \ref{Table3} summarizes the normalized mean score \cite{score} of each optimizer over all sklearn models (lower the better). The score is normalized to (-1, 1) w.r.t the random search where 1 means random search level performance and 0 means the optimum found by random search \cite{score}. We make two observations:
\begin{itemize}
\item Overall $turbo$ is the best optimizer and $Random Search$ is the worst optimizer, in terms of  number of models with which the optimizer achieves the lowest loss. 
\item Each optimizer is good at different models, which also presents a chance of an ensemble of optimizers.
\end{itemize}

\begin{table}
  \caption{Average optimization time breakdown per iteration: seconds. The budget is 40 seconds per iteration.}
  \label{Table2}
  \centering
  \begin{tabular}{lllllll}
    \cmidrule(r){1-7}
     & hyperopt & Random Search & opentuner & pysot & skopt & turbo \\
    \midrule
    suggest & 0.008 & 0.009 & 0.038 & 0.026 & 0.345 & 0.385    \\
    observe & 0.00006 & 0.009 & 0.005 & 0.001 & 0.185 & 0.004
  \end{tabular}
\end{table}

\begin{table}
  \caption{Average minimum loss of each optimizer over all scikit-learn models. The lowest loss from each model (per row) is highlighted.}
  \label{Table3}
  \centering
  \begin{tabular}{lllllll}
    \cmidrule(r){1-7}
     & hyperopt & Random Search & opentuner & pysot & skopt & turbo\\
    \midrule
    DT & 0.049 & 0.161 & 0.11 & -0.01 & -0.015 & \textcolor{blue}{\textbf{-0.007}} \\
    MLP-adam & 0.011 & 0.023 & 0.039 & 0.016 & 0.037 & \textcolor{blue}{\textbf{-0.007}} \\
    MLP-sgd & 0.035 & 0.059 & 0.084 & 0.016 & 0.026 & \textcolor{blue}{\textbf{-0.008}} \\
    RF & 0.006 & 0.033 & 0.081 & -0.02 & \textcolor{blue}{\textbf{-0.024}} & 0.007 \\
    SVM & 0.035 & 0.074 & 0.039 & 0.025 & 0.035 & \textcolor{blue}{\textbf{0.012}} \\
    ada & 0.132 & 0.125 & \textcolor{blue}{\textbf{0.069}} & 0.101 & 0.104 & 0.099 \\
    kNN & 0.022 & 0.065 & \textcolor{blue}{\textbf{0.0}} & 0.047 & 0.056 & 0.045 \\
    lasso & \textcolor{blue}{\textbf{0.027}} & 0.068 & 0.03 & 0.041 & 0.062 & 0.034 \\
    linear & 0.012 & 0.06 & 0.065 & \textcolor{blue}{\textbf{-0.001}} & 0.022 & 0.013 \\
  \end{tabular}
\end{table}

\section{A Heuristic Ensemble Algorithm for Optimizers.}
We propose a heuristic algorithm for an ensemble of two optimizers, which can be easily generalized to multiple optimizers. In Algorithm 1 (Right), two optimizers $opt\_1$ and $opt\_2$ are initialized. At each iteration, the optimizer is supposed to suggest $N\_BATCH$ guesses. Instead, $opt\_1$ and $opt\_2$ each contribute half of the $N\_BATCH$ guesses. A key design choice is that when the evaluation scores return, they are passed to both optimizers so that the two optimizers can learn from each other’s suggestions.

\begin{center}
\begin{minipage}[t]{6.8cm}
  \vspace{0pt}  
  \begin{algorithm}[H]
   opt = Some\_Optimizer()\\
 \For{iter\_id = 1 to N\_STEP} {
  params\_list = opt.suggest(N\_BATCH)\\
  scores = evaluate(params\_list)\\
  opt.observe(scores)\\
 }
  \end{algorithm}
\end{minipage}%
\begin{minipage}[t]{6.8cm}
  \vspace{0pt}
  \begin{algorithm}[H]
    opt\_1 = Some\_Optimizer()\\
    opt\_2 = Another\_Optimizer()\\
 \For{iter\_id = 1 to N\_STEP} {
  params\_list\_1 = opt1.suggest(N\_BATCH/2)\\
  params\_list\_2 = opt2.suggest(N\_BATCH/2)\\
  \# concatenate two lists \\ 
  params\_list = params\_list\_1 + params\_list\_2 \\
  scores = evaluate(params\_list)\\
  opt\_1.observe(scores)\\
  opt\_2.observe(scores)\\
 }
  \end{algorithm}
\end{minipage}
\end{center}

\begin{center}
   Algorithm 1: Left: workflow of a single optimizer. Right: A heuristic ensemble algorithm for two optimizers. 
\end{center}

\section{GPU Accelerated Exhaustive Search}
A key question is which two optimizers to choose for the ensemble. Based on the motivation study in Table 3, we believe that an exhaustive search of all possible combinations is the most reliable method. 
However, using the default models and datasets of Bayesmark has the following downsides:

\begin{itemize}
  \item The data size is small. The number of samples of built-in toy dataset ranges from 150 to 1797. Small dataset introduces randomness and it is prone to overfitting.
  \item Evaluating scikit-learn models on the CPU is slow. For example, the total running time to obtain the results of Table 3 is 42 hours using the default scripts. It should be noted that this experiment is performed on the small sklearn toy data. Running additional data will be even more time consuming.
\end{itemize}

To make the experiments more representative and robust, We add three new real-world datasets: California housing, hotel booking and Higgs Boson. Each dataset is down-sampled to 10000 samples to make experiments faster.

\begin{figure}[h!]
\centering
\includegraphics[width=0.85\textwidth]{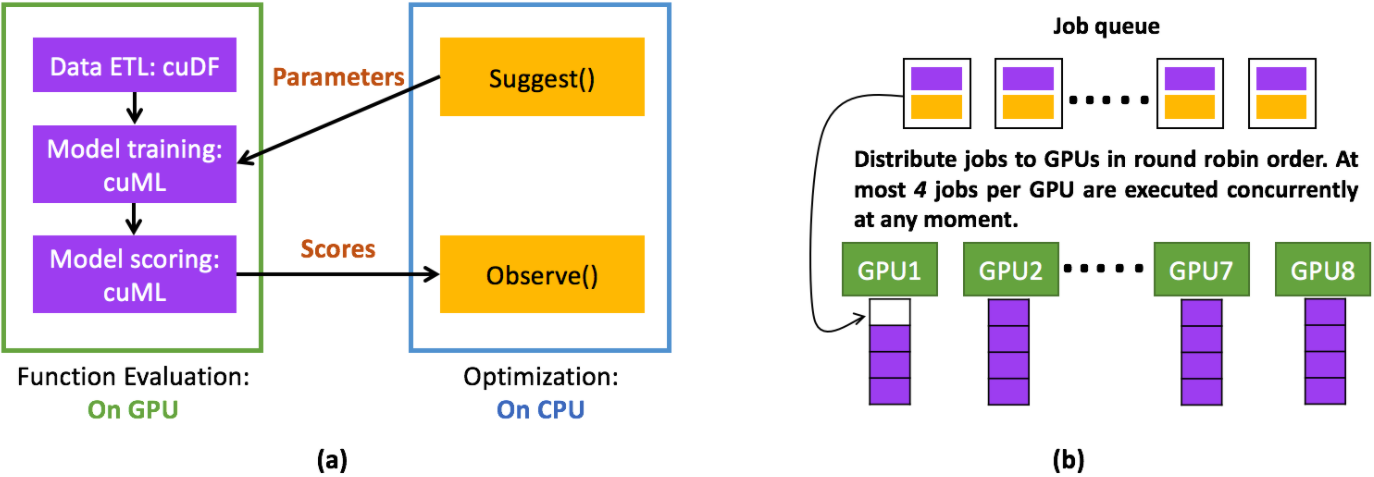}
\caption{GPU acceleration of BBO. (a) GPUs are used to execute computing intensive function evaluations with cuDF and cuML libraries. (b) Parallel execution of function evaluation and optimization on multiple GPUs.} 
\label{fig:method1}
\end{figure}

To accelerate the experiments, we implement the entire evaluation pipeline on GPU utilizing the RAPIDS GPU data science framework. Data loading and preprocessing are boosted by cuDF \cite{cudf} while scikit-learn models and scorers are replaced with their GPU counterparts in cuML library \cite{raschka2020machine}. Xgboost \cite{Chen:2016:XST:2939672.2939785}  is also added to the benchmark suite, which supports both GPU and CPU modes. We also implemented a GPU-optimized MLP using pytorch \cite{NEURIPS2019_9015} where data loader is implemented with cupy \cite{nishino2017cupy} and DLPACK \cite{dlpack} to keep data on GPU. 

Another benefit of moving function evaluation onto GPUs is workload partitioning as shown in Figure 1(a). The more computing intensive workloads are on GPU while the relatively lightweight optimization is on the CPU. This also facilitates parallel experiments when multiple GPUs are available. The workloads are distributed to each GPU in round robin order until each GPU has $N$ jobs to execute concurrently. For Bayesmark with cuML workloads, we experiment with several values for $N$ and set $N=4$ for optimal GPU utilization and memory consumption.

\section{Experiments and Results}
\subsection{Hardware Configuration}
The experiments are performed on a DGX-1 workstation with 8 NVIDIA V100 GPUs and two 20-core Intel Xeon CPUs. An exhaustive search is implemented for all $\binom{M}{2}$ combinations of $M$ provided optimizers. In this case, $M=5$ so there are 10 ensembles and 5 single optimizers to experiment. We exclude $Random Search$ because it has the worst performance and crashes randomly in ensemble for GPU estimators \cite{bayesmark}.

\subsection{Evaluation of the Ensemble of Optimizers}
\begin{figure}
\centering
\includegraphics[width=0.95\textwidth]{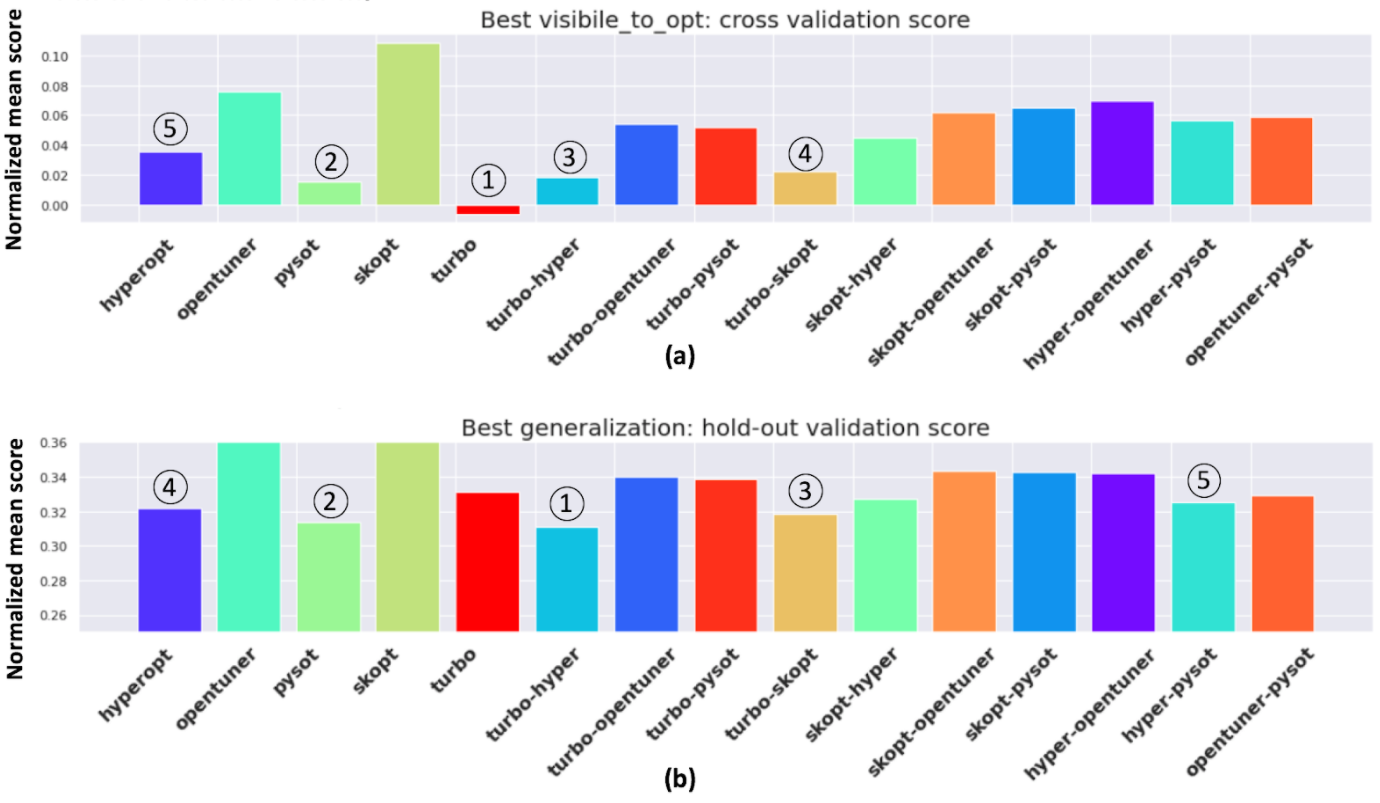}
\caption{Performance of optimization algorithms in terms of (a) cross validation score that is visible to and minimized by optimizers and (b) holdout validation score which represents the generalization ability of the optimizer. The y-axis is normalized mean score \cite{score} and lower is better. The top 5 optimizers are highlighted in each sub-figure}
\label{fig:method2}
\end{figure}

Figure 2 summarizes the performance of each optimization algorithm. In Bayesmark, the data is split into training data and hold-out test data. The cross validation score of the training data is visible to optimizers and it is the score optimized. The validation score of the hold-out test data represents the generalization capability of the optimizers to new data. We argue that the generalization score is more important because the hidden models and dataset in the competition must be different. For example, $turbo$ has some degree of overfitting since it is the best optimizer for cross validation but it is not in top 5 in terms of generalization. Based on these results, we believe the best three optimizers overall are $turbo-hyperopt$, $pysot$ and $turbo-skopt$.

\begin{figure}
\centering
\includegraphics[width=0.95\textwidth]{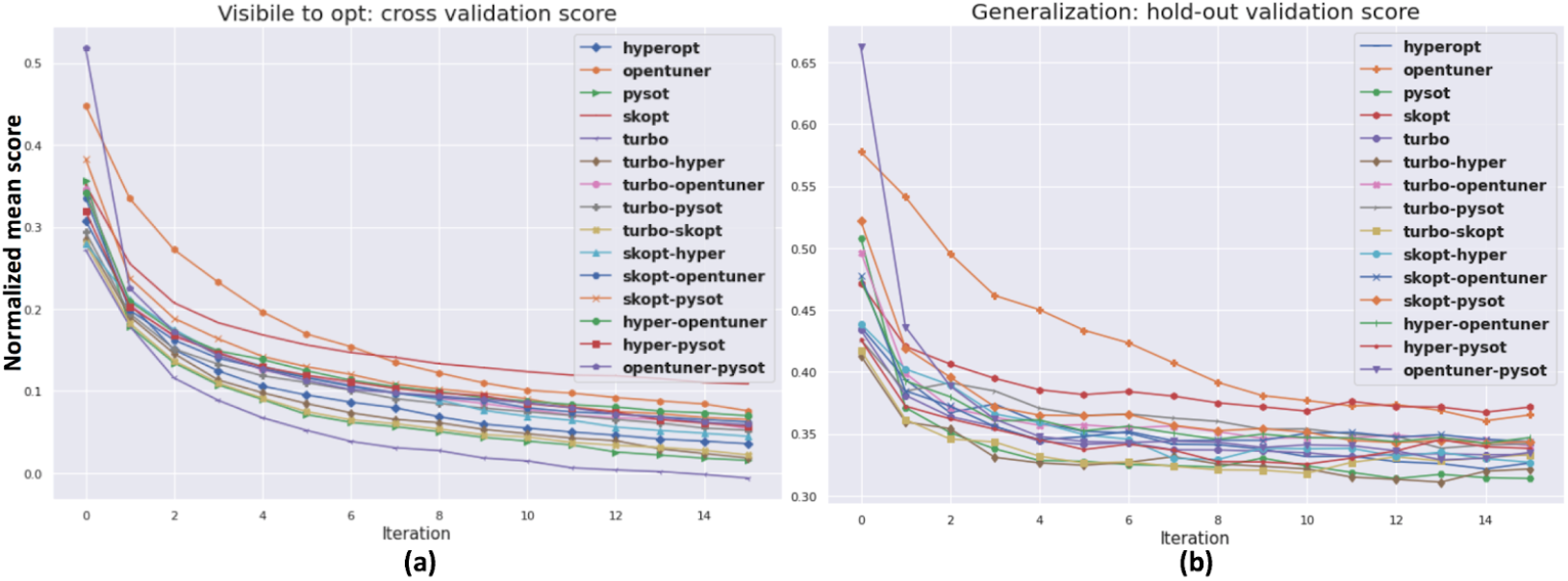}
\caption{Performance of optimization algorithms per iteration: (a) cross validation score that is visible to optimizers. (b) holdout validation score which represents the generalization ability.}
\label{fig:method3}
\end{figure}

Figure \ref{fig:method3} shows the iterative performance of each optimization algorithm. Since the cross validation score is visible to the optimizers and the cumulative minimum is used, the curve always goes down. It is clear that optimizer $turbo$ (purple) outperforms every other optimizer by a large margin as shown in Figure \ref{fig:method3}(a). However, Figure \ref{fig:method3}(b) shows a different pattern. The $turbo-hyper$ (brown diamond) and $turbo-skopt$ (yellow square) converge faster than other optimizers including the best single optimizer $pysot$ (green). We believe that it is due to the diversity of ensemble compared to a single optimizer.

\begin{figure}
\centering
\includegraphics[width=0.99\textwidth]{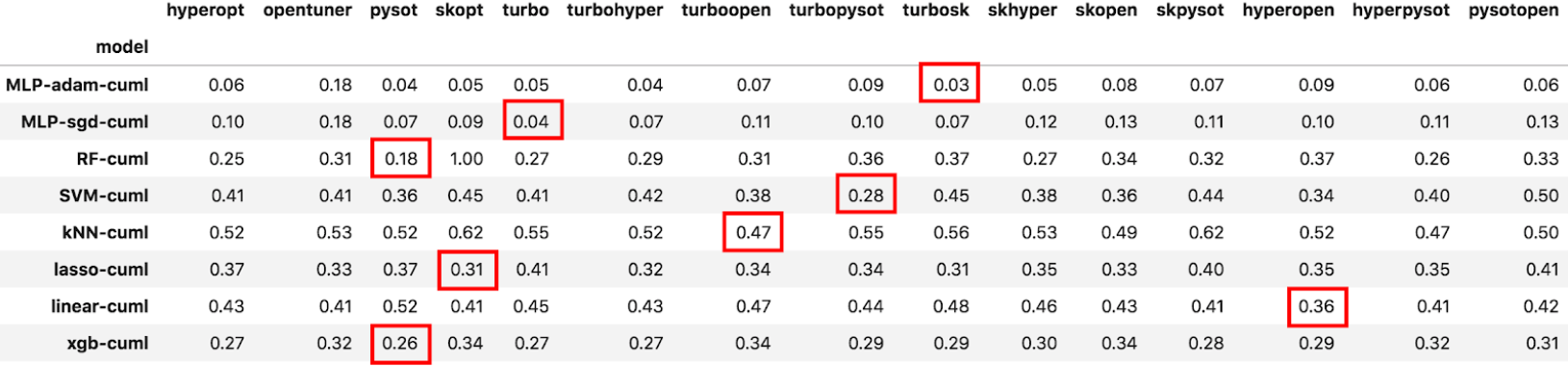}
\caption{Generalization performance of optimizers on each cuML model. The best optimizer for each model (per row) is highlighted.}
\vspace{0.5 cm}
\label{fig:method4}
\end{figure}

Figure \ref{fig:method4} shows the performance breakdown of each optimizer on each of the cuML models, in terms of the generalization score. The optimizer $pysot$ has the best performance for two tree based models: random forest and xgboost. The ensemble optimizer $turbo-skopt$ shines at $MLP-adam$, the most widely used deep learning model in the benchmark suite. It is also interesting to note that the best ensemble optimizer $turbo-hyper$ does not achieve the best performance for any model particularly.

Based on these results, we believe the best three optimizers overall are $turbo-hyperopt$, $pysot$ and $turbo-skopt$, which can be good submissions to the competition. \textbf{We chose $turbo-skopt$ as our final submission because 1) it has a Top-3 generalization score; 2) it converges faster than single optimizers and 3) it achieves best performance for a representative deep learning model. Using this simple ensemble, we won the $2^{nd}$ place of NeurIPS 2020 black-box optimization challenge.}

\begin{table}
  \caption{Final Submission. Higher leaderboard score is better.}
  \label{final}
  \centering
  \begin{tabular}{llll}
    \cmidrule(r){1-4}
     & turbo & skopt & turboskopt \\
    \midrule
    Final leaderboard ranking & $36^{th}$ & $24^{th}$ & \textcolor{green}{$2^{nd}$} \\
    Final leaderboard score (0-100) & 88.08 & 88.9 & \textcolor{green}{92.9} \\
    Local generalization score (0-100) & 64 & 66.5 & \textcolor{green}{68.1} \\
    
  \end{tabular}
  \vspace{0.5 cm}
\end{table}

Table~\ref{final} summarizes the performance of our final submission $turbo-skopt$. Our ensemble method (LB 92.9 ranking $2^{nd}$) improves significantly upon the single optimizers it consists of, namely $turbo$ (LB 88.9 ranking $24^{th}$) and $skopt$ (LB 88.08 ranking $36^{th}$) on the final leaderboard. Similar improvement is also evident in our local validation.

\begin{figure}
\centering
\includegraphics[width=0.95\textwidth]{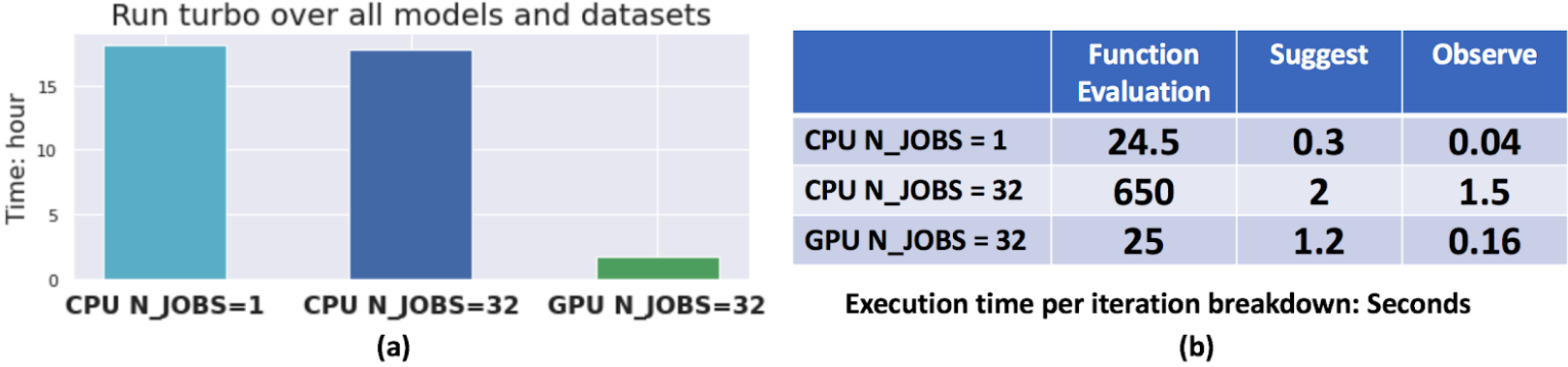}
\caption{(a) Running time comparison between the proposed multi-GPU implementation and multi-core CPU implementation. (b) The breakdown of execution time per iteration.}
\label{fig:method5}
\end{figure}

\subsection{GPU Speedup}
The proposed multi-GPU implementation is the key to finish the exhaustive search in reasonable time. We define a job as one invocation of “bayesmark.experiment()” or one “bayesmark-launch” command with one model, one dataset, one metric and one optimizer. As shown in Figure 5(a), it takes 1.8 hours on GPUs to examine the optimizer turbo over all models, datasets and metrics, where 32 jobs run concurrently on GPUs. In contrast, the multi-core CPUs can’t take advantage of parallelism as shown in Figure 5(a) due to  two reasons:

\begin{itemize}
\item Some models utilize multi-thread training by default such as xgboost and MLP. The multiple cores of CPUs are already busy even when running with one job at a time. Figure 5(b) shows that the function evaluation time of CPU with $N\_JOBS=32$ almost scales linearly with respect to CPU with $N\_JOBS=1$, indicating there is no speedup.
\item The optimizers also run on CPUs so both models and optimizers are competing for CPU resources and slow down the overall performance.
\end{itemize}

In contrast, the multi-GPU implementation naturally isolates the workloads: models are on GPUs while optimizers are on CPUs. Each GPU is also isolated without any contention with other GPUs. Overall the proposed multi-GPU implementation achieves 10x speedup of the CPU counterpart and all the experiments (Figure 2, 3 and 4) \textbf{finish in 24 hours which consists of 4,230 jobs, 2.7 million model trainings and 541,440 optimizations.} The same workload would take at least 10 days on CPUs. 

\begin{figure}
\centering
\includegraphics[width=0.99\textwidth]{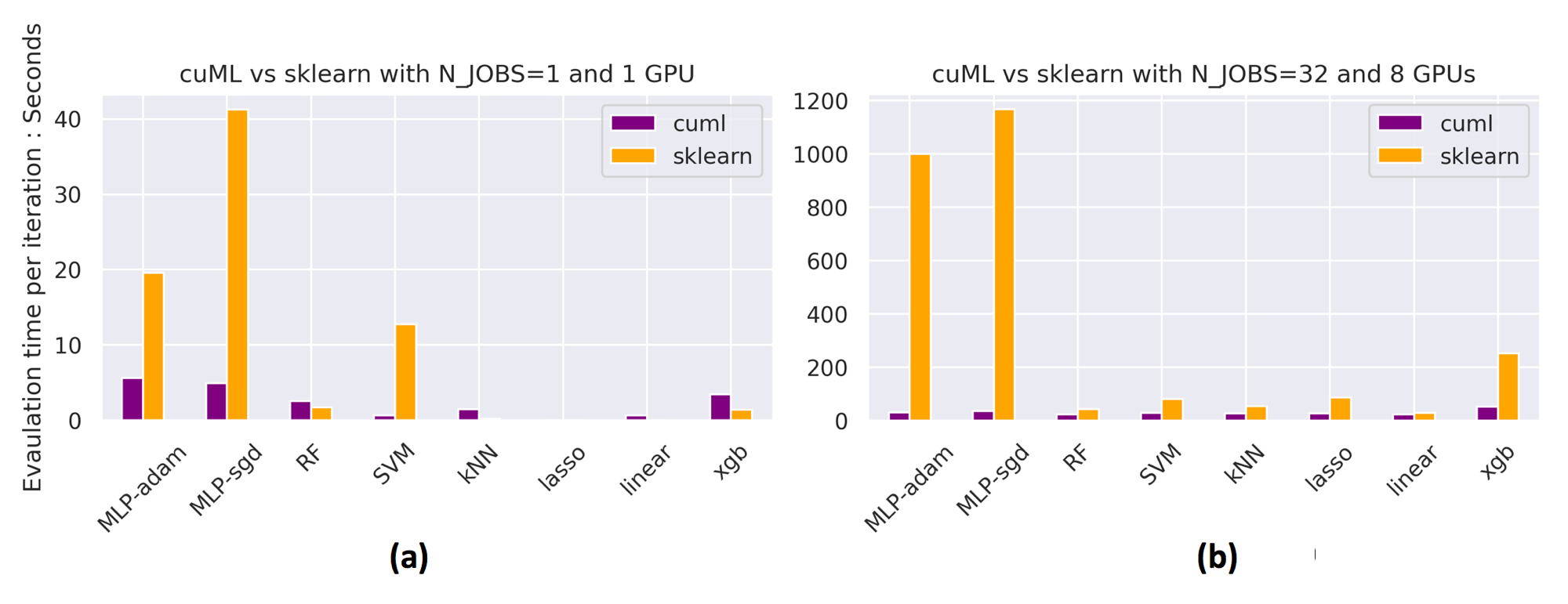}
\caption{Run time comparison of cuML vs sklearn models.}
\label{fig:time}
\end{figure}

Figure~\ref{fig:time} shows the detailed run time comparison of cuML models and sklearn models. Since the dataset in this experimentation is small, cuML models such as $knn$ and $xgb$ could be slower than their sklearn counterparts on CPU when only one job is training. When launching many jobs, cuML achieves significant speedup on all of the models.

\section{Conclusion}
In this paper, we propose a fast multi-GPU accelerated exhaustive search to find the best ensemble of optimization algorithms. The ensemble algorithm can be generalized to multiple optimizers and the proposed framework also scales with multiple GPUs.

\bibliographystyle{unsrt}
\bibliography{ref}

\begin{thebibliography}{10}

\bibitem{golovin2017google}
Daniel Golovin, Benjamin Solnik, Subhodeep Moitra, Greg Kochanski, John Karro,
  and D~Sculley.
\newblock Google vizier: A service for black-box optimization.
\newblock In {\em Proceedings of the 23rd ACM SIGKDD international conference
  on knowledge discovery and data mining}, pages 1487--1495, 2017.

\bibitem{eriksson2019scalable}
David Eriksson, Michael Pearce, Jacob Gardner, Ryan~D Turner, and Matthias
  Poloczek.
\newblock Scalable global optimization via local bayesian optimization.
\newblock In {\em Advances in Neural Information Processing Systems}, pages
  5496--5507, 2019.

\bibitem{bayesmark}
Uber.
\newblock uber/bayesmark.
\newblock {https://github.com/uber/bayesmark}.

\bibitem{bergstra2012random}
James Bergstra and Yoshua Bengio.
\newblock Random search for hyper-parameter optimization.
\newblock {\em The Journal of Machine Learning Research}, 13(1):281--305, 2012.

\bibitem{shahriari2015taking}
Bobak Shahriari, Kevin Swersky, Ziyu Wang, Ryan~P Adams, and Nando De~Freitas.
\newblock Taking the human out of the loop: A review of bayesian optimization.
\newblock {\em Proceedings of the IEEE}, 104(1):148--175, 2015.

\bibitem{hernandez2017parallel}
Jos{\'e}~Miguel Hern{\'a}ndez-Lobato, James Requeima, Edward~O Pyzer-Knapp, and
  Al{\'a}n Aspuru-Guzik.
\newblock Parallel and distributed thompson sampling for large-scale
  accelerated exploration of chemical space.
\newblock {\em arXiv preprint arXiv:1706.01825}, 2017.

\bibitem{optuna-2019}
Takuya Akiba, Shotaro Sano, Toshihiko Yanase, Takeru Ohta, and Masanori Koyama.
\newblock Optuna: A next-generation hyperparameter optimization framework.
\newblock In {\em Proceedings of the 25rd {ACM} {SIGKDD} International
  Conference on Knowledge Discovery and Data Mining}, 2019.

\bibitem{bergstra2013making}
James Bergstra, Daniel Yamins, and David Cox.
\newblock Making a science of model search: Hyperparameter optimization in
  hundreds of dimensions for vision architectures.
\newblock In {\em International conference on machine learning}, pages
  115--123. PMLR, 2013.

\bibitem{bergstra2011algorithms}
James~S Bergstra, R{\'e}mi Bardenet, Yoshua Bengio, and Bal{\'a}zs K{\'e}gl.
\newblock Algorithms for hyper-parameter optimization.
\newblock In {\em Advances in neural information processing systems}, pages
  2546--2554, 2011.

\bibitem{blackbox}
Find the best black-box optimizer for machine learning.
\newblock {https://bbochallenge.com/}.

\bibitem{raschka2020machine}
Sebastian Raschka, Joshua Patterson, and Corey Nolet.
\newblock Machine learning in python: Main developments and technology trends
  in data science, machine learning, and artificial intelligence.
\newblock {\em arXiv preprint arXiv:2002.04803}, 2020.

\bibitem{Chen:2016:XST:2939672.2939785}
Tianqi Chen and Carlos Guestrin.
\newblock {XGBoost}: A scalable tree boosting system.
\newblock In {\em Proceedings of the 22nd ACM SIGKDD International Conference
  on Knowledge Discovery and Data Mining}, KDD '16, pages 785--794, New York,
  NY, USA, 2016. ACM.

\bibitem{scikit-learn}
F.~Pedregosa, G.~Varoquaux, A.~Gramfort, V.~Michel, B.~Thirion, O.~Grisel,
  M.~Blondel, P.~Prettenhofer, R.~Weiss, V.~Dubourg, J.~Vanderplas, A.~Passos,
  D.~Cournapeau, M.~Brucher, M.~Perrot, and E.~Duchesnay.
\newblock Scikit-learn: Machine learning in {P}ython.
\newblock {\em Journal of Machine Learning Research}, 12:2825--2830, 2011.

\bibitem{ansel:pact:2014}
Jason Ansel, Shoaib Kamil, Kalyan Veeramachaneni, Jonathan Ragan-Kelley,
  Jeffrey Bosboom, Una-May O'Reilly, and Saman Amarasinghe.
\newblock Opentuner: An extensible framework for program autotuning.
\newblock In {\em International Conference on Parallel Architectures and
  Compilation Techniques}, Edmonton, Canada, Aug 2014.

\bibitem{eriksson2019pysot}
David Eriksson, David Bindel, and Christine~A Shoemaker.
\newblock pysot and poap: An event-driven asynchronous framework for surrogate
  optimization.
\newblock {\em arXiv preprint arXiv:1908.00420}, 2019.

\bibitem{skopt}
Scikit-optimize: Sequential model-based optimization in python.
\newblock {https://scikit-optimize.github.io/stable/}.

\bibitem{datametrics}
bayesmark data and metrics.
\newblock {https://github.com/uber/bayesmark\#launch-the-experiments}.

\bibitem{runlocal}
bbo challenge starter kit.
\newblock
  {https://github.com/rdturnermtl/bbo\_challenge\_starter\_kit/blob/master/run\_local.sh}.

\bibitem{score}
uber/bayesmark.
\newblock How scoring works.
\newblock
  {https://bayesmark.readthedocs.io/en/latest/scoring.html\#mean-scores}.

\bibitem{cudf}
cudf - gpu dataframes.
\newblock {https://github.com/rapidsai/cudf}.

\bibitem{NEURIPS2019_9015}
Adam Paszke, Sam Gross, Francisco Massa, Adam Lerer, James Bradbury, Gregory
  Chanan, Trevor Killeen, Zeming Lin, Natalia Gimelshein, Luca Antiga, Alban
  Desmaison, Andreas Kopf, Edward Yang, Zachary DeVito, Martin Raison, Alykhan
  Tejani, Sasank Chilamkurthy, Benoit Steiner, Lu~Fang, Junjie Bai, and Soumith
  Chintala.
\newblock Pytorch: An imperative style, high-performance deep learning library.
\newblock In H.~Wallach, H.~Larochelle, A.~Beygelzimer, F.~d~\textquotesingle
  Alch\'{e}-Buc, E.~Fox, and R.~Garnett, editors, {\em Advances in Neural
  Information Processing Systems 32}, pages 8024--8035. Curran Associates,
  Inc., 2019.

\bibitem{nishino2017cupy}
ROYUD Nishino and Shohei Hido~Crissman Loomis.
\newblock Cupy: A numpy-compatible library for nvidia gpu calculations.

\bibitem{dlpack}
Dlpack: Open in memory tensor structure.
\newblock {https://github.com/dmlc/dlpack}.

\end{thebibliography}

\end{document}